\documentclass[10pt,twocolumn,letterpaper]{article}

\usepackage{cvpr}
\usepackage{times}
\usepackage{epsfig}
\usepackage{graphicx}
\usepackage{amsmath}
\usepackage{amssymb}
\usepackage[dvipsnames]{xcolor}
\usepackage[numbers,sort,compress]{natbib}
\usepackage{booktabs}
\usepackage{bbding}
\usepackage{multirow, makecell}
\usepackage{titletoc}

\definecolor{mypink}{RGB}{219, 48, 122}

\usepackage[pagebackref=true,breaklinks=true,letterpaper=true,colorlinks,bookmarks=false]{hyperref}
    
\cvprfinalcopy %

\def\cvprPaperID{2935} %

\ifcvprfinal\fi

\begin{document}

\title{Leveraging Photometric Consistency over Time for \\ Sparsely Supervised Hand-Object Reconstruction}
\author{
 Yana Hasson\textsuperscript{1,2} \thanks{This work was performed during an internship at Microsoft.}
 \qquad Bugra Tekin\textsuperscript{4}
 \qquad Federica Bogo\textsuperscript{4}\\
 \qquad Ivan Laptev\textsuperscript{1,2}
 \qquad Marc Pollefeys\textsuperscript{3,4}
 \qquad Cordelia Schmid\textsuperscript{1,5}
 \\ \\
 {\normalsize \textsuperscript{1}Inria, \textsuperscript{2}D\'{e}partement d'informatique de l'ENS, CNRS, PSL Research University} \\
 {\normalsize   \textsuperscript{3}ETH Z\"{u}rich, \textsuperscript{4}Microsoft},
 {\normalsize \textsuperscript{5}Univ. Grenoble Alpes, CNRS, Grenoble INP, LJK}
}

\maketitle
\thispagestyle{empty}

\begin{abstract}
		
Modeling hand-object manipulations is essential for {understanding} how humans interact with their environment.
While of practical importance, estimating the pose of hands and objects during interactions is challenging due to the large mutual occlusions that occur during manipulation.
Recent efforts have been directed towards fully-supervised methods that require large amounts of labeled training samples. 
Collecting 3D ground-truth data for hand-object interactions, however, is costly, tedious, and error-prone. 
To overcome this challenge we present a method to leverage photometric consistency across time when annotations are only available for a sparse subset of frames in a video.
Our model is trained end-to-end on color images to jointly reconstruct hands and objects in 3D by inferring their poses.
Given our estimated reconstructions, we differentiably render the optical flow between pairs of adjacent images and use it within the network to warp one frame to another.
We then apply a self-supervised photometric loss that relies on the visual consistency between nearby images.
We achieve state-of-the-art results on 3D hand-object reconstruction benchmarks and demonstrate that our approach allows us to improve the pose estimation accuracy by leveraging information from neighboring frames in low-data regimes.
\end{abstract}

\section{Introduction}

Understanding how hands interact with objects is crucial for a semantically meaningful interpretation of human action and behavior.
In recent years, impressive hand pose estimation results have been demonstrated, but joint prediction of hand and object poses has received so far only limited attention, although unified 3D modeling of hands and objects is essential for many applications in augmented reality, robotics and surveillance. 

\begin{figure}[t]
	\begin{center}
   \includegraphics[width=1.0\linewidth]{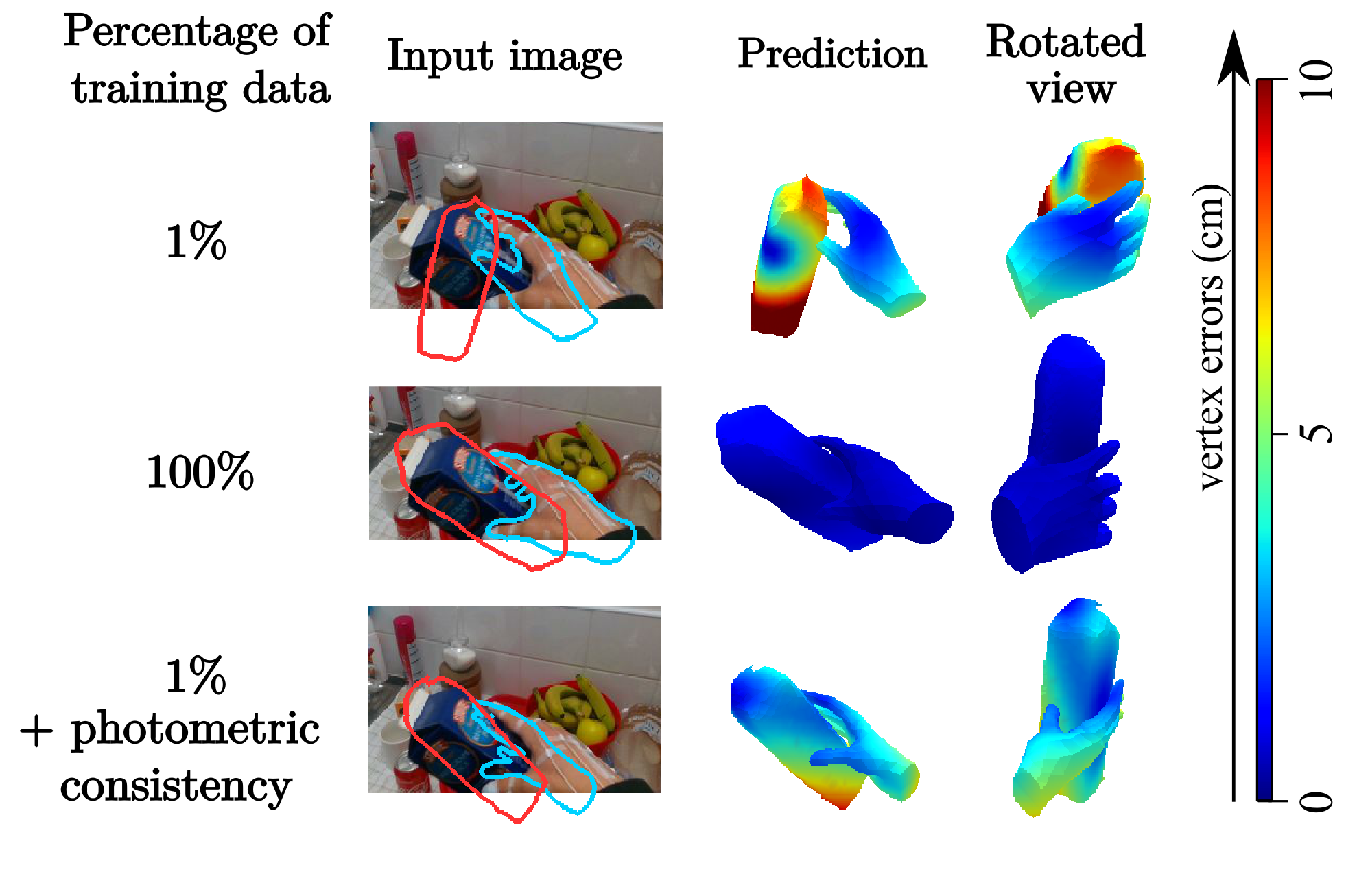}
\mbox{}\vspace{-0.4cm}\\
	\end{center}
	\caption{Our method provides accurate 3D hand-object reconstructions from monocular, sparsely annotated RGB videos.
	We introduce a loss which exploits photometric consistency between neighboring frames. The loss effectively propagates information from a few annotated frames to the rest of the video.}
\mbox{}\vspace{-1.0cm}\\
	\label{fig:teaser}
\end{figure}

Estimating the pose of hands during interaction with an object is an extremely challenging problem due to mutual occlusions. 
Joint 3D reconstruction of hands and objects is even more challenging as this would require precise understanding of the subtle interactions that take place in cluttered real-world environments. 
Recent work in computer vision has been able to tackle some of the challenges in unified understanding of hands and objects for color input. 
Pioneering works of~\cite{iqbal2018ECCV,GANeratedHands_CVPR2018,romero2010b} have proposed ways to recover hand motion during object manipulation, yet without explicitly reasoning about the object pose. 
Recent few efforts to model hand-object interactions~\cite{hasson19_obman,tekin19_handplusobject}, on the other hand, have focused on joint 3D hand-object pose estimation and reconstruction techniques. 
However, these methods require full-supervision on large datasets with 3D hand-object pose annotations.  
Collecting such 3D ground-truth datasets for hand-object interactions remains a challenging problem. 

While motion capture datasets~\cite{hernando2018cvpr} can provide large amounts of training samples with accurate annotations, they can only be captured in controlled settings and have visible markers on the images that bias pose prediction in color images.
Multi-view setups~\cite{Simon_2017_CVPR,Zimmermann_2019_ICCV}, which enable 3D triangulation from 2D detections, can similarly only be captured in controlled environments.
Synthetic datasets provide an alternative. However, existing ones~\cite{hasson19_obman,GANeratedHands_CVPR2018,Mueller2017ICCV,zb2017hand} cannot yet reach the fidelity and realism to generalize to real datasets.
Manual annotation and optimization-based techniques for data annotation can be slow and error-prone.
Due to these challenges associated with data collection, existing datasets are either real ones that are limited in size and confined to constrained environments or synthetic ones that lack realism.
Models trained on such data are more prone to overfitting and lack generalization capabilities.

Our method aims at tackling these challenges and reduces the stringent reliance on 3D annotations.
To this end, we propose a novel weakly supervised approach to joint 3D hand-object reconstruction.
Our model jointly estimates the hand and object pose and reconstructs their shape in 3D, given training videos with annotations in only sparse frames on a small fraction of the dataset.
Our method  models the temporal nature of 3D hand and object interactions and leverages motion as a self-supervisory signal for 3D dense hand-object reconstruction.
An example result is shown in Fig.~\ref{fig:teaser}.

Our contributions can be summarized as follows:
\begin{itemize}
	\item{We present a new method for joint dense reconstruction of hands and objects in 3D. Our method operates on color images and efficiently regresses model-based shape and pose parameters in a single feed-forward pass through a neural network.}
	\item{We introduce a novel photometric loss that relies on the estimated optical flow between pairs of adjacent images. Our scheme leverages optical flow to warp one frame to the next, directly within the network, and exploits the visual consistency between successive warped images with a self-supervised loss, ultimately alleviating the need for strong supervision.}
\end{itemize}

In Section~\ref{sec:eval}, we show quantitatively that these contributions  allow  us  to  reliably predict the pose of interacting hands and objects in 3D, while densely reconstructing their 3D shape. Our approach allows us to improve  pose estimation accuracy in the absence of strong supervision on challenging real-world sequences and achieves state-of-the-art results on 3D hand-object reconstruction benchmarks.
The code is publicly available.~\footnote{\url{https://hassony2.github.io/handobjectconsist}}
\section{Related Work}
\label{sec:related}

Our work tackles the problem of estimating hand-object pose from monocular RGB videos, exploiting photometric cues for self-supervision.
To the best of our knowledge, our method is the first to apply such self-supervision to hand-object scenarios.
We first review the literature on hand and object pose estimation.
Then, we focus on methods using motion and photometric cues for self-supervision, in particular in the context of human body pose estimation.

\paragraph{Hand and object pose estimation.}
Most approaches in the literature tackle the problem of estimating either hand or object pose, separately.

For object pose estimation from RGB images, the recent trend is to use convolutional neural networks (CNNs) to predict the 2D locations of the object's 3D bounding box in image space~\cite{Kehl2017,Rad2017,tekin18}.
The 6D pose is then obtained via PnP~\cite{pnp} or further iterative refinement.
Such methods commonly need a 3D model of the object as input, and large amounts of labeled data.
DeepIM~\cite{li2018deepim} shows generalization to unseen objects by iteratively matching rendered images of an object against observed ones.
Recently, Pix2Pose~\cite{pix2pose} improves robustness against occlusions by predicting dense 2D-3D correspondences between image pixels and the object model.
Most methods~\cite{li2018deepim,pix2pose,sundermeyer2018} try to limit the amount of required annotations by relying on synthetic data. 
However, it remains unclear how well these methods would perform in the presence of large occlusions as the ones caused by hand-object interactions.

Several approaches for hand pose estimation from RGB images focus on regressing 3D skeleton joint positions~\cite{cai2018_weakly,iqbal2018ECCV,GANeratedHands_CVPR2018,spurr2018cvpr,yang2019,zb2017hand}.
However, methods that directly output 3D hand surfaces offer a richer representation, and allow one to directly reason about occlusions and contact points~\cite{mueller2019}.
Parametric hand models like MANO~\cite{MANO:SIGGRAPHASIA:2017} represent realistic 3D hand meshes using a set of shape and pose parameters.
~\cite{Totalcap2018,pavlakos2018humanshape} fit such parametric models to CNN-regressed 2D joint positions to estimate hand poses from full-body images.

A number of recent methods plug MANO into end-to-end deep learning frameworks, obtaining accurate hand 3D shape and pose from single RGB images~\cite{Boukhayma_2019_CVPR,Ge2019,Zhang_2019_ICCV}.
Similarly to us, these approaches regress the model parameters directly from the image, though they do not address scenarios with hand-object interactions. 
Given the challenges involved, hand-object interactions have been tackled in multi-view or RGB-D camera setups~\cite{RogezSR15,Tzionas:IJCV:2016}.
Targeting pose estimation from single RGB images, Romero et al.~\cite{romero2010b} obtain 3D hand-object reconstructions via nearest neighbor search in a large database of synthetic images.

Recently, efforts have been put into the acquisition of ground-truth 3D annotations for both hands and objects during interaction.
Early datasets which provide annotated RGB views of hands manipulating objects rely on manual annotations~\cite{RealtimeHO_ECCV2016} and depth tracking~\cite{Tzionas:IJCV:2016}, which limits the size and the occlusions between hand and object.
Larger datasets which rely on motion capture~\cite{hernando2018cvpr} and multi-view setups~\cite{hampali2019cvpr} have been collected, spurring the development of new methods for hand-object pose estimation.
Recently, \cite{hasson19_obman,tekin19_handplusobject} propose CNN-based approaches to accurately predict hand and object poses from monocular RGB.
However, these methods are fully supervised and do not exploit the temporal dimension for pose estimation.

\paragraph{Supervision using motion and photometric cues.}

In RGB videos, motion cues provide useful information that can be used for self-supervision.
Several methods explore this idea in the context of human body pose estimation.

Pfister et al.~\cite{pfister2015flowing} leverage optical flow for 2D human pose estimation.
Slim DensePose~\cite{neverova19slim} uses an off-the-shelf optical flow method~\cite{flownet2} to establish dense correspondence~\cite{densepose} between adjacent frames in a video.
These correspondences are used to propagate manual annotations between frames and to enforce spatio-temporal equivariance constraints. 
Very recently, PoseWarper~\cite{bertasius2019neurips} leverages image features to learn the pose warping between a labeled frame and an unlabeled one, thus propagating annotations in sparsely labeled videos.

Regressing 3D poses is more difficult: the problem is fundamentally ambiguous in monocular scenarios. Furthermore, collecting 3D annotations is not as easy as in 2D.
VideoPose3D~\cite{pavllo:videopose3d:2019} regresses 3D skeleton joint positions, by back-projecting them on the image space and using CNN-estimated 2D keypoints as supervision.
Tung et al.~\cite{tung2017self} regress the SMPL body model parameters~\cite{SMPL:2015} by using optical flow and reprojected masks as weak supervision.
Differently from us, they rely on an off-the-shelf optical flow method, making the pose accuracy dependent on the flow quality.
Recently, Arnab et al.~\cite{Arnab_CVPR_2019} refine noisy per-frame pose predictions~\cite{hmrKanazawa18} using bundle adjustment over the SMPL parameters.
These methods are not tested in scenarios with large body occlusions. 

Our method enforces photometric consistency between pose estimates from adjacent frames. 
Similar ideas have been successfully applied to self-supervised learning of ego-motion, depth and scene flow for self-driving cars~\cite{Brickwedde_2019_ICCV,monodepth17,zhoucvpr2017}.
Unlike these methods, which estimate pixel-wise probability depth distributions for mostly rigid scenes, we focus on estimating the articulated pose of hands manipulating objects.
Starting from multi-view setups at training time, \cite{Rhodin_eccv2018,Rhodin_cvpr2018} propose weak supervision strategies for monocular human pose estimation.
We consider monocular setups where the camera might move. %
Similarly to us, TexturePose~\cite{Pavlakos_2019_ICCV} enforces photometric consistency between pairs of frames to refine body pose estimates.
They define the consistency loss in UV space: this assumes a UV parameterization is always provided. Instead, we define our loss in image space. 
Notably, these methods consider scenarios without severe occlusions (only one instance, \ie one body, is in the scene).

None of these methods focuses on hands, and more particularly on complex hand-object interactions.

\section{Method}

We propose a CNN-based model for 3D hand-object reconstruction that can be efficiently trained from a set of \emph{sparsely annotated} video frames.
Namely, our method takes as input a monocular RGB video, capturing hands interacting with objects.
We assume that the object model is known, and that sparse annotations are available only for a subset of video frames.

As in previous work~\cite{tekin18,tekin19_handplusobject,Kehl2017}, we assume that a 3D mesh model of the object is provided.
To reconstruct hands, we rely on the parametric model MANO~\cite{MANO:SIGGRAPHASIA:2017}, which deforms a 3D hand mesh template according to a set of shape and pose parameters. 
As output, our method returns hand and object 3D vertex locations (together with shape and pose parameters) for each frame in the sequence.

The key idea of our approach is to use a photometric consistency loss, that we leverage as self-supervision on the unannotated intermediate frames in order to improve hand-object reconstructions.
We introduce this loss in Sec.~\ref{sec:photometric}. We then describe our learning framework in detail in Sec.~\ref{sec:network}.

\begin{figure*}
	\begin{center}
	   \includegraphics[width=0.98\linewidth]{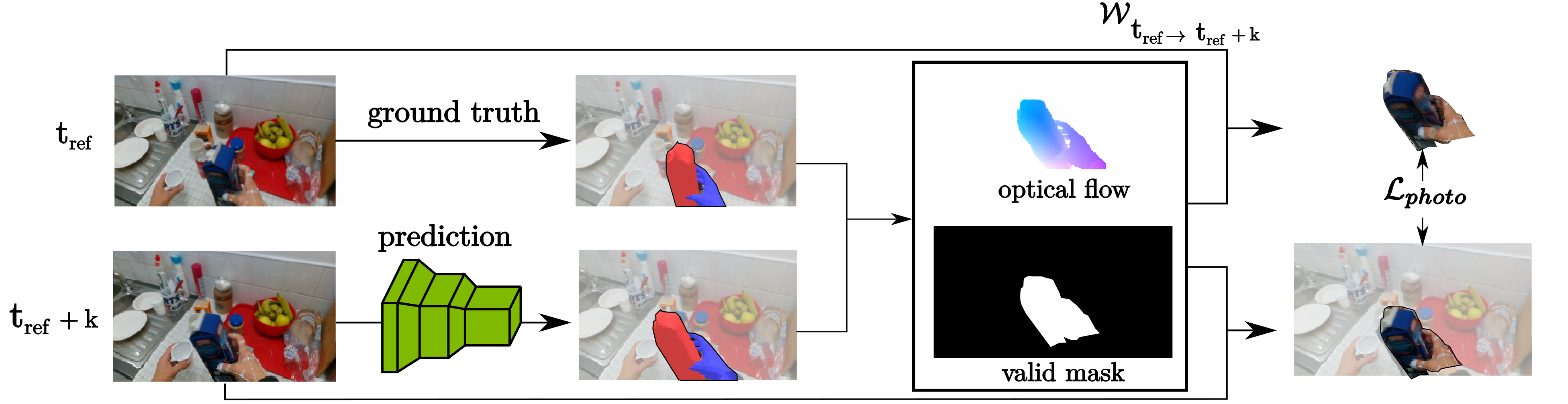}
\mbox{}\vspace{-0.2cm}\\
	\end{center}{}
  \caption{Photometric consistency loss. Given an annotated frame, $t_{ref}$, and an unannotated one, $t_{{ref}+k}$, we reconstruct hand and object 3D pose at $t_{{ref}+k}$ leveraging a self-supervised loss.
  We differentiably render the optical flow between ground-truth hand-object vertices at $t_{ref}$ and estimated ones.
  Then, we use this flow to warp frame $t_{{ref}+k}$ into $t_{ref}$, and enforce consistency in pixel space between warped and real image.}
  \mbox{}\vspace{-1.2cm}\\
	\label{fig:consist}
\end{figure*}

\begin{figure}
  \begin{center}
     \includegraphics[width=0.98\linewidth]{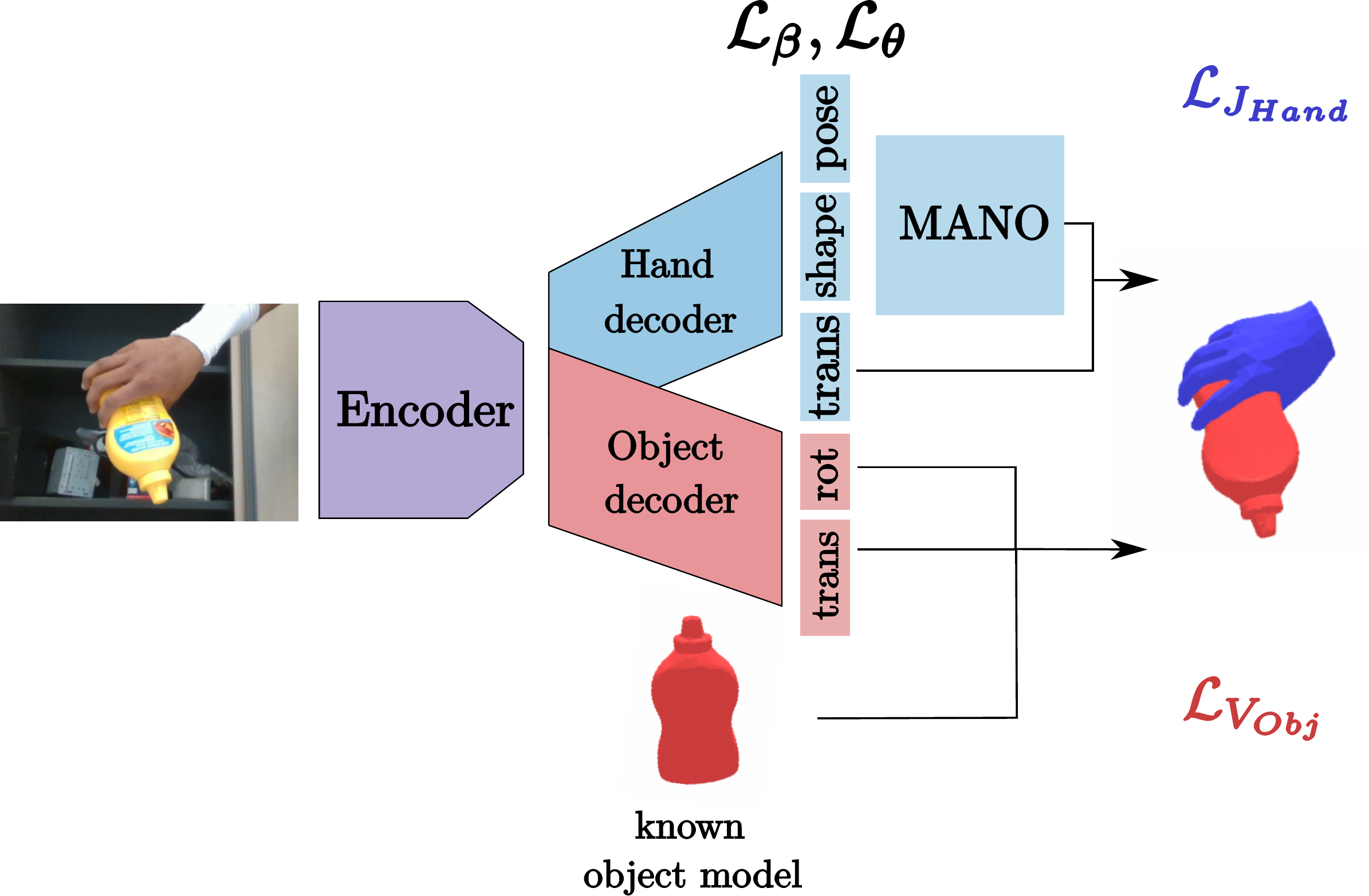}
  \end{center}
  \caption{Architecture of the single-frame hand-object reconstruction network.}
  \mbox{}\vspace{-0.8cm}\\
  \label{fig:architecture}
  \mbox{}\vspace{-1.0cm}\\
\end{figure}

\subsection{Photometric Supervision from Motion}
\label{sec:photometric}

As mentioned above, our method takes as input a sequence of RGB frames and outputs hand and object mesh vertex locations for each frame.
The same type of output is generated in~\cite{hasson19_obman}, where each RGB frame is processed separately.
We observe that the temporal continuity in videos imposes temporal constraints between neighboring frames.
We assume that 3D annotations are provided only for a sparse subset of frames; this is a scenario that often occurs in practice when data collection is performed on sequential images,
but only a subset of them is manually annotated. We then define a self-supervised loss to propagate this information to unlabeled frames.

Our self-supervised loss exploits photometric consistency between frames, and is defined in image space. Figure~\ref{fig:consist} illustrates the process.
Consider an annotated frame at time $t_{ref}$, $I_{t_{ref}}$, for which we have ground-truth hand and object vertices $V_{t_{ref}}$ (to simplify the notation, we do not distinguish here between hand and object vertices).
Given an unlabeled frame $I_{t_{ref}+k}$, our goal is to accurately regress hand and object vertex locations $V_{t_{ref}+k}$.
Our main insight is that, given estimated per-frame 3D meshes and known camera intrinsics, we can back-project our meshes on image space and leverage pixel-level information to provide additional cross-frame supervision.

Given $I_{t_{ref}+k}$, we first regress hand and object vertices $V_{t_{ref}+k}$ in a single feed-forward network pass (see Sec.~\ref{sec:network}).
Imagine now to back-project these vertices on $I_{t_{ref}+k}$ and assign to each vertex the color of the pixel they are projected onto.
The object meshes at $t_{ref}$ and $t_{{ref}+k}$ share the same topology; and so do the hand meshes.
So, if we back-project the ground-truth meshes at $t_{ref}$ on $I_{t_{ref}}$, corresponding vertices from $V_{t_{ref}}$ and $V_{t_{ref}+k}$ should be assigned the same color.

We translate this idea into our photometric consistency loss.
We compute the 3D displacement (``flow'') between corresponding vertices from $V_{t_{ref}}$ and $V_{t_{ref}+k}$.
These values are then projected on the image plane, and interpolated on the visible mesh triangles. 
To this end, we differentiably render the estimated flow from $V_{t_{ref}}$ to $V_{t_{ref}+k}$ using the Neural Renderer~\cite{kato2018renderer}.
This allows us to define a warping flow $W$ between the pair of images as a function of $V_{t_{ref}+k}$.

We exploit the computed flow to warp $I_{t_{ref}+k}$ into the warped image $\mathcal{W}(I_{t_{ref}+k}, V_{t_{ref}+k})$, by differentiably sampling values from $I_{t_{ref}+k}$ according to the predicted optical flow displacements.
Our loss enforces consistency between the warped image and the reference one:
\begin{equation} 
  \mathcal{L}_{photo}(V_{t_{ref}+k}) = ||M \cdot (\mathcal{W}(I_{t_{ref}+k}, V_{t_{ref}+k}) - I_{t_{ref}})||_1,
\end{equation}
where $M$ is a binary mask denoting surface point visibility.
In order to compute the visibility mask, we ensure that the supervised pixels belong to the silhouette of the reprojected mesh in the target frame $I_{{t_{ref}}+k}$.
We additionally verify that the supervision is not applied to pixels which are occluded in the reference frame by performing a cyclic consistency check similarly to~\cite{Hur_2017_ICCV,neverova19slim} which is detailed in Appendix~\ref{sec:cyclic}.
We successively warp a grid of pixel locations using the optical flow $t_{ref}$ to $t_{ref}+k$ and from $t_{ref} + k$ to $t_{ref}$ and include only pixel locations which remain stable, a constraint which does not hold for mesh surface points which are occluded in one of the frames.
Note that the error is minimized with respect to the estimated hand and object vertices $V_{t_{ref}+k}$.

The consistency supervision $\mathcal{L}_{photo}$ can be applied directly on pixels, similarly to self-supervised ego-motion and depth learning scenarios~\cite{zhoucvpr2017,monodepth17}.
The main difference with these approaches is that they estimate per-pixel depth values while we attempt to leverage the photometric consistency loss in order to refine rigid and articulated motions.
Our approach is similar in spirit to that of~\cite{Pavlakos_2019_ICCV}. With respect to them, we consider a more challenging scenario (multiple 3D instances and large occlusions).
Furthermore, we define our loss in image space, instead of UV space, and thus we do not assume that a UV parametrization is available.

As each operation is differentiable, we can combine this loss and use it as supervision either in isolation or in addition to other reconstruction losses (Sec.~\ref{sec:network}).

\subsection{Dense 3D Hand-Object Reconstruction}
\label{sec:network}

We apply the loss introduced in Sec.~\ref{sec:photometric} to 3D hand-object reconstructions obtained independently for each frame.
These per-frame estimates are obtained with a single forward pass through a deep neural network, whose architecture is shown in Fig.~\ref{fig:architecture}.
In the spirit of~\cite{hasson19_obman,Boukhayma_2019_CVPR}, our network takes as input a single RGB image and outputs MANO~\cite{MANO:SIGGRAPHASIA:2017} pose and shape parameters.
However, differently from~\cite{hasson19_obman}, we assume that a 3D model of the object is given, and we regress its 6D pose by adding a second head to our network (see again Fig.~\ref{fig:architecture}).
We employ as backbone a simple ResNet-18~\cite{He2015}, which is computationally very efficient (see Sec.~\ref{sec:eval}).
We use the base network model as the image encoder and select the last layer before the classifier to produce our image features.
We then regress hand and object parameters from these features through $2$ dense layers with ReLU non-linearities. Further details about the architecture can be found in Appendix~\ref{sec:implemdetails}.

In the following, we provide more details about hand-object pose and shape regression, and about the losses used at training time.

\textbf{Hand-object global pose estimation.} We formulate the hand-object global pose estimation problem in the camera coordinate system and aim to find precise absolute 3D positions of hands and objects. 
Instead of a weak perspective camera model, commonly used in the body pose estimation literature, we choose here to use a more realistic projective model.
In our images, hand-object interactions are usually captured at a short distance from the camera. So the assumptions underlying weak perspective models do not hold.
Instead, we follow best practices from object pose estimation.
As in~\cite{li2018deepim,xiang2017posecnn}, we predict values that can be easily estimated from image evidence.
Namely, in order to estimate hand and object translation,
we regress a focal-normalized depth offset $d_f$ and a 2D translation vector $(t_u,t_v)$, defined in pixel space.
We compute $d_f$ as
\begin{equation}
  d_f = \frac{V_z - z_{off}}{f},
\end{equation}
\noindent where $V_z$ is the distance between mesh vertex and camera center along the z-axis, $f$ is the camera focal length, and $z_{off}$ is empirically set to $40cm$.
$t_u$ and $t_v$ represent the translation, in pixels, of the object (or hand) origin, projected on the image space, with respect to the image center.
Note that we regress $d_f$ and $(t_u,t_v)$ for both the hand and the object, separately.

Given the estimated $d_f$ and $(t_u,t_v)$, and the camera intrinsics parameters, we can easily derive the object (hand) global translation in 3D.
For the global rotation, we adopt the axis-angle representation.
Following~\cite{hmrKanazawa18,li2018deepim,pavlakos2018humanshape}, the rotation for object and hand is predicted in the object-centered coordinate system.

\textbf{Hand articulated pose and shape estimation.}
We obtain hand 3D reconstructions by predicting MANO pose and shape parameters.
For the pose, similarly to~\cite{hasson19_obman,Boukhayma_2019_CVPR}, we predict the principal composant analysis (PCA) coefficients of the low-dimensional hand pose space provided in~\cite{MANO:SIGGRAPHASIA:2017}.
For the shape, we predict the MANO shape parameters, which control identity-specific characteristics such as skeleton bone length.
Overall, we predict $15$ pose coefficients and $10$ shape parameters. %

\textbf{Regularization losses.} We find it effective to regularize both hand pose and shape by applying $\ell_2$ penalization as in~\cite{Boukhayma_2019_CVPR}.
$\mathcal{L}_{\theta_{\mathit{Hand}}}$ prevents unnatural joint rotations, while $\mathcal{L}_{\beta_{\mathit{Hand}}}$ prevents extreme shape deformations, which can result in irregular and unrealistic hand meshes.

\textbf{Skeleton adaptation.} Hand skeleton models can vary substantially between datasets, resulting in inconsistencies in the definition of joint locations.
Skeleton mismatches may force unnatural deformations of the hand model.
To account for these differences, we replace the fixed MANO joint regressor with a skeleton adaptation layer which regresses joint locations from vertex positions.
We initialize this linear regressor using the values from the MANO joint regressor and optimize it jointly with the network weights. 
We keep the tips of the fingers and the wrist joint fixed to the original locations, and learn a dataset-specific mapping for the other joints at training time. More details are provided in Appendix~\ref{sec:skeletonadapt}.

\textbf{Reconstruction losses.} 
In total, we predict $6$ parameters for hand-object rotation and translation and $25$ MANO parameters, which result in a total of $37$ regressed parameters.
We then apply the predicted  transformations to the reference hand and object models and further produce the 3D joint locations of the MANO hand model, which are output by MANO in addition to the hand vertex locations.
We define our supervision on hand joint positions, $\mathcal{L}_{J_{\mathit{Hand}}}$, as well as on 3D object vertices, $\mathcal{L}_{V_{\mathit{Obj}}}$. 
Both losses are defined as $\ell_2$ errors.

Our final loss $\mathcal{L}_{HO}$ is a weighted sum of the reconstruction and regularization terms:
\begin{equation}
\mathcal{L}_{HO} = \mathcal{L}_{V_{\mathit{Obj}}}
    +  \lambda_{J} \mathcal{L}_{J_{\mathit{Hand}}} 
    + \lambda_{\beta} \mathcal{L}_{\beta_{\mathit{Hand}}}
    +  \lambda_{\theta} \mathcal{L}_{\theta_{\mathit{Hand}}}.
\end{equation}

\section{Evaluation}
\label{sec:eval}

In this section, we first describe the datasets and corresponding evaluation protocols. We then compare our method to the state of the art and provide a detailed analysis of our framework.

\subsection{Datasets} \label{subsec:datasets}

We evaluate our framework for joint 3D hand-object reconstruction and pose estimation on two recently released datasets: First Person Hand Action Benchmark~\cite{hernando2018cvpr} and HO-3D~\cite{hampali2019cvpr} which provide pose annotations for all hand keypoints as well as the manipulated rigid object.
 
\textbf{First-person hand action benchmark (FPHAB):} The FPHAB dataset~\cite{hernando2018cvpr} collects egocentric RGB-D videos capturing a wide range of hand-object interactions, with ground-truth annotations for 3D hand pose, 6D object pose, and hand joint locations. 
The annotations are obtained in an automated way, using mocap magnetic sensors strapped on hands.
Object pose annotations are available for $4$ objects, for a subset of the videos. 
Similarly to hand annotations, they are obtained via magnetic sensors.
In our evaluation, we use the same \emph{action split} as in~\cite{tekin19_handplusobject}: each object is present in both the training and test splits, thus allowing the model to learn instance-specific 6 degrees of freedom (DoF) transformations. To further compare our results to those of~\cite{hasson19_obman}, we also use the \emph{subject split} of FPHAB where the training and test splits feature different subjects.

\textbf{HO-3D:} The recent HO-3D dataset~\cite{hampali2019cvpr} is the result of an effort to collect 3D pose annotations for both hands and manipulated objects in a markerless setting.
In this work, we report results on the subset of the dataset which was released as the first version~\cite{hampali2019ho3d}.
Details on the specific subset are provided in Appendix~\ref{sec:ho3dsub}.
The subset of HO-3D we focus on contains $14$ sequences, out of which $2$ are available for evaluation.
The authors augment the real training sequences with additional synthetic data.
In order to compare our method against the baselines introduced in~\cite{hampali2019ho3d}, we train jointly on their real and synthetic training sets.

\begin{figure*}

    \begin{center}
       \includegraphics[width=0.95\linewidth]{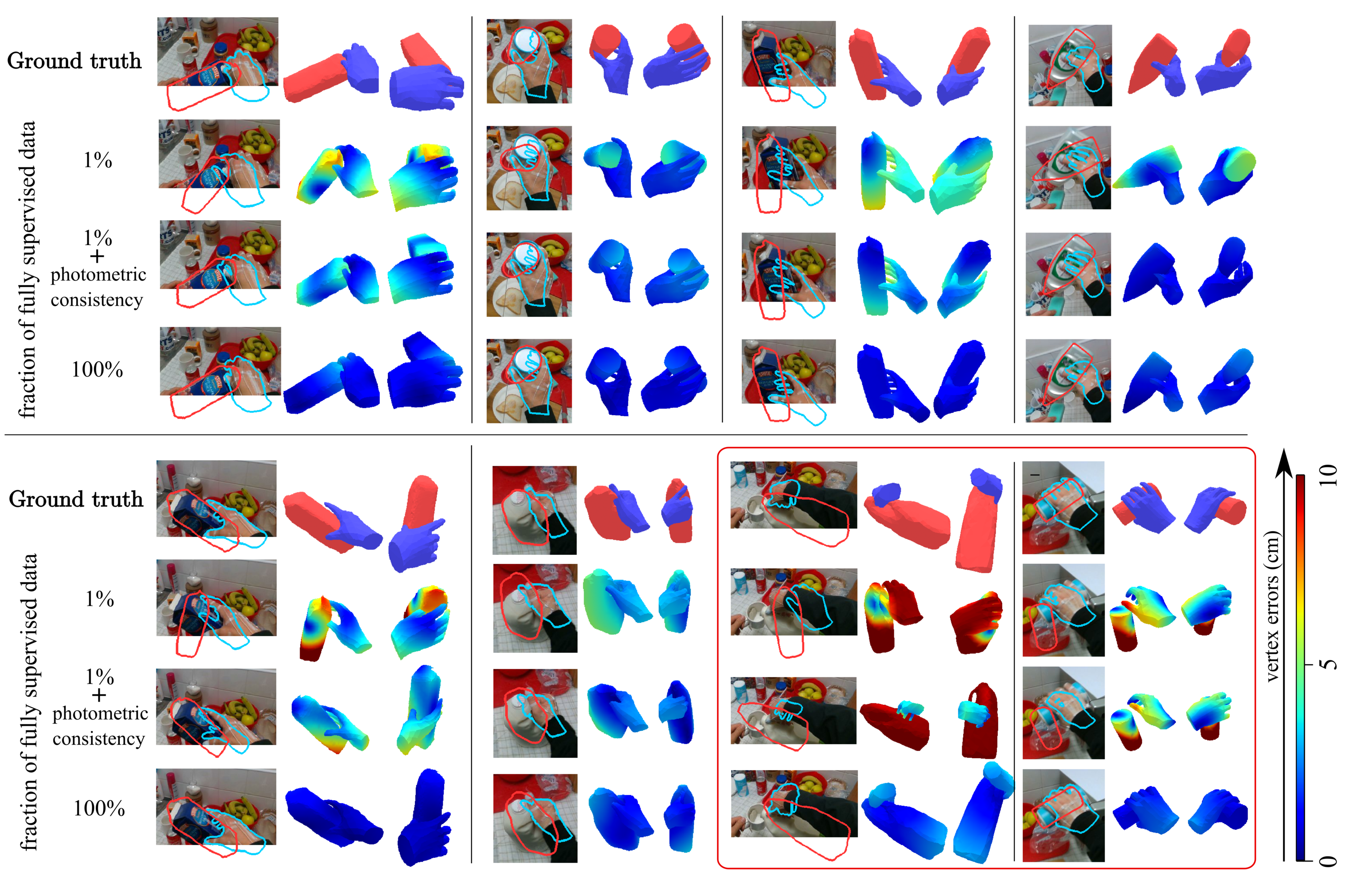}
    \end{center}{}
    \mbox{}\vspace{-1.0cm}\\
\caption{Qualitative results on the FPHAB dataset. We visualize the reconstructed meshes reprojected on the image as well as a rotated view. When training on the full dataset, we obtain reconstructions which accurately capture the hand-object interaction. In the sparsely supervised setting, we qualitatively observe that photometric consistency allows to recover more accurate hand and object poses. Failure cases occur in the presence of important motion blur and large occlusions of the hand or the object by the subject's arm.}
    \mbox{}\vspace{-0.8cm}\\
    \label{fig:quali_fhb}
\end{figure*}

\subsection{Evaluation Metrics} \label{subsec:metrics}

We evaluate our approach on 3D hand pose estimation and 6D object pose estimation and use official train/test splits to evaluate our performance in comparison to the state of the art. 
We report accuracy using the following metrics.

\textbf{Mean 3D errors.} To assess the quality of our 3D hand reconstructions, we compute the mean end-point error (in mm) over $21$ joints following~\cite{zb2017hand}. 
For objects, on FPHAB we compute the average vertex distance (in mm) in camera coordinates to compare against~\cite{tekin19_handplusobject}, on HO-3D, we look at average bounding box corner distances.

\textbf{Mean 2D errors.} We report the mean errors between reprojected keypoints and 2D ground-truth locations for hands and objects.
To evaluate hand pose estimation accuracy, we measure the average joint distance. 
For object pose estimation, following the protocol for 3D error metrics, we report average 2D vertex distance on FPHAB, and average 2D corner distance on HO-3D.
To further compare our results against~\cite{hampali2019ho3d}, we also report the percentage of correct keypoints (PCK).
To do so, for different pixel distances, we compute the percentage of frames for which the average error is lower than the given threshold.

\subsection{Experimental Results}

We first report the pose estimation accuracy of our single-frame hand-object reconstruction model and compare it against the state of the art~\cite{hampali2019ho3d,tekin19_handplusobject}.
We then present the results of our motion-based self-supervised learning approach and demonstrate its efficiency in case of scarcity of ground-truth annotations.

\begin{table}
	\begin{center}
		\begin{tabular}{|l|c|c|}
			\hline
			Method & Hand error & Object error \\
			\hline\hline
			Tekin \etal  & \textbf{15.8} & 24.9 \\
			Ours & 18.0 & \textbf{22.3} \\
			\hline
		\end{tabular}
	\end{center}
	\caption{Comparison to state-of-the-art method of Tekin \etal \cite{tekin19_handplusobject} on FPHAB~\cite{hernando2018cvpr}, errors are reported in mm.}
	\label{table:comp_tekin}
    \mbox{}\vspace{-0.5cm}\\
\end{table}

\begin{table}
	\begin{center}
		\begin{tabular}{|l|c|}
			\hline
			Method  &  Hand error \\
			\hline\hline
			Ours - no skeleton adaptation & 28.1  \\
			Ours & \textbf{27.4}  \\
			Hasson \etal~\cite{hasson19_obman} & 28.0 \\
			\hline
		\end{tabular}
	\end{center}
	\caption{On the FHPAB dataset, for which the skeleton is substantially different from the MANO one, we show that adding a skeleton adaptation layer allows us to outperform~\cite{hasson19_obman}, while additionally predicting the global translation of the hand.} %
\mbox{}\vspace{-1.0cm}\\
\end{table}

\textbf{Single-frame hand-object reconstruction.}
Taking color images as input, our model reconstructs dense meshes to leverage pixel-level consistency, and infers hand and object poses.
To compare our results to the state of the art~\cite{hampali2019ho3d,hasson19_obman,tekin19_handplusobject}, we evaluate our pose estimation accuracy on the FPHAB~\cite{hernando2018cvpr} and HO-3D~\cite{hampali2019cvpr} datasets.

Table~\ref{table:comp_tekin} demonstrates that our model achieves better accuracy than~\cite{tekin19_handplusobject} on object pose estimation.
We attribute this to the fact that~\cite{tekin19_handplusobject} regresses keypoint positions, and recovers the object pose as a non-differentiable post-processing step, while we directly optimize for the 6D pose.
Our method achieves on average a hand pose estimation error of $18$ mm on FPHAB which is outperformed by~\cite{tekin19_handplusobject} by a margin of $2.6$ mm.
This experiment is in line with earlier reported results, where the estimation of individual keypoint locations outperformed regression of model parameters~\cite{Pavlakos_2019_ICCV,hmrKanazawa18,pavlakos2018humanshape}.
While providing competitive pose estimation accuracy to the state of the art, our approach has the advantage of predicting a detailed hand shape, which is crucial for fine-grained understanding of hand-object interactions and contact points. We further compare our results to those of~\cite{hasson19_obman} that reports results on FPHAB using the \emph{subject split} and demonstrate that our model provides improved hand pose estimation accuracy, while additionally estimating the global position of the hand in the camera space.

\begin{figure}[t]
\mbox{}\vspace{-0.5cm}\\
	\begin{center}
		\vspace{-3mm}
		\includegraphics[width=1.0\linewidth]{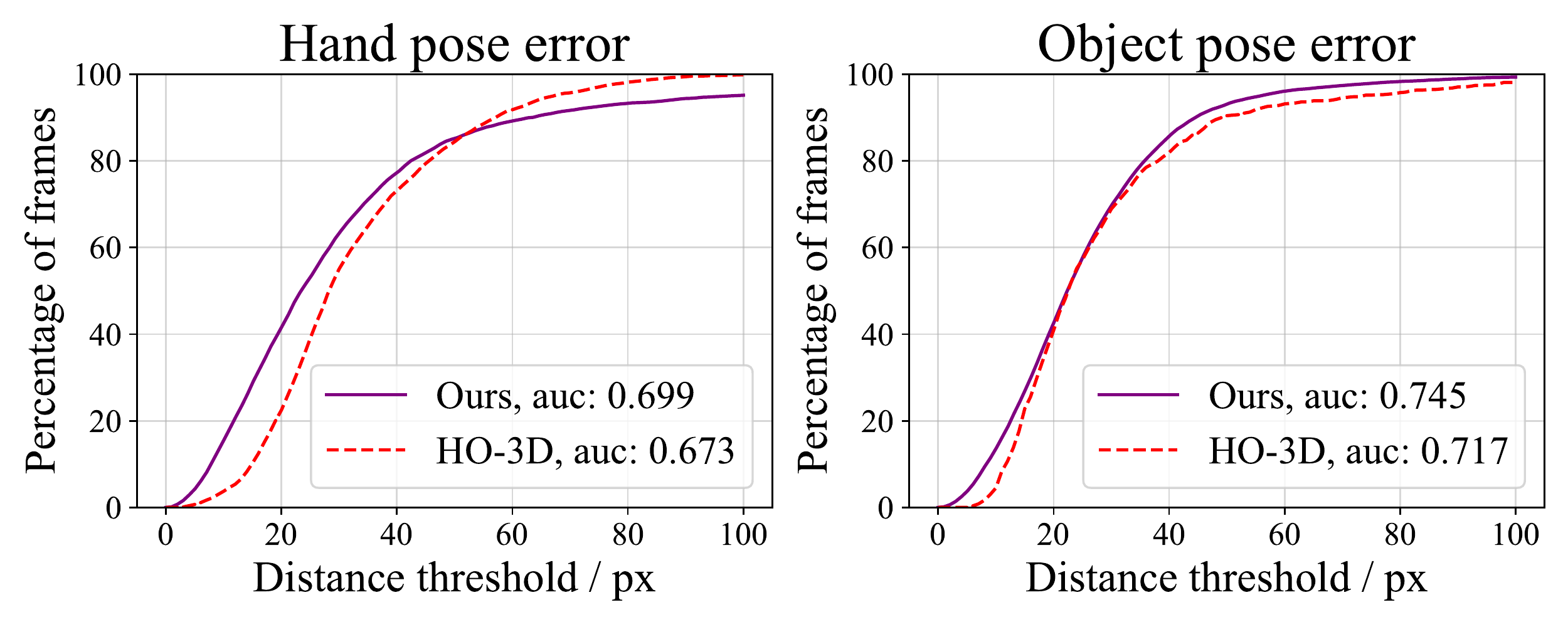}
	\end{center}
	\caption{Evaluation of our baseline for hand-object pose estimation on the early release of the HO-3D~\cite{hampali2019ho3d} dataset. We report the PCK for 2D joint mean-end-point error for hands, and the mean 2D reprojection error for objects.}
    \mbox{}\vspace{-0.6cm}\\
	\label{fig:compareho3d}
\end{figure}

We further evaluate the hand-object pose estimation accuracy of our single-image model on the recently introduced HO-3D dataset. 
We show in Fig.~\ref{fig:compareho3d} that we outperform~\cite{hampali2019ho3d} on both hand and object pose estimation. 

In Table~\ref{table:sharenc}, we analyze the effect of simultaneously training for hand and object pose estimation within a unified framework. We compare the results of our unified model to those of the models trained individually for hand pose estimation and object pose estimation. We observe that the unified co-training slightly degrades hand pose accuracy. This phenomenon is also observed by~\cite{tekin19_handplusobject}, and might be due to the fact that while the hand pose highly constrains the object pose, simultaneous estimation of the object pose does not result in increased hand pose estimation accuracy, due to higher degrees of freedom inherent to the articulated pose estimation problem.

\begin{figure}
\mbox{}\vspace{-.4cm}\\
    \begin{center}
       \includegraphics[width=0.98\linewidth]{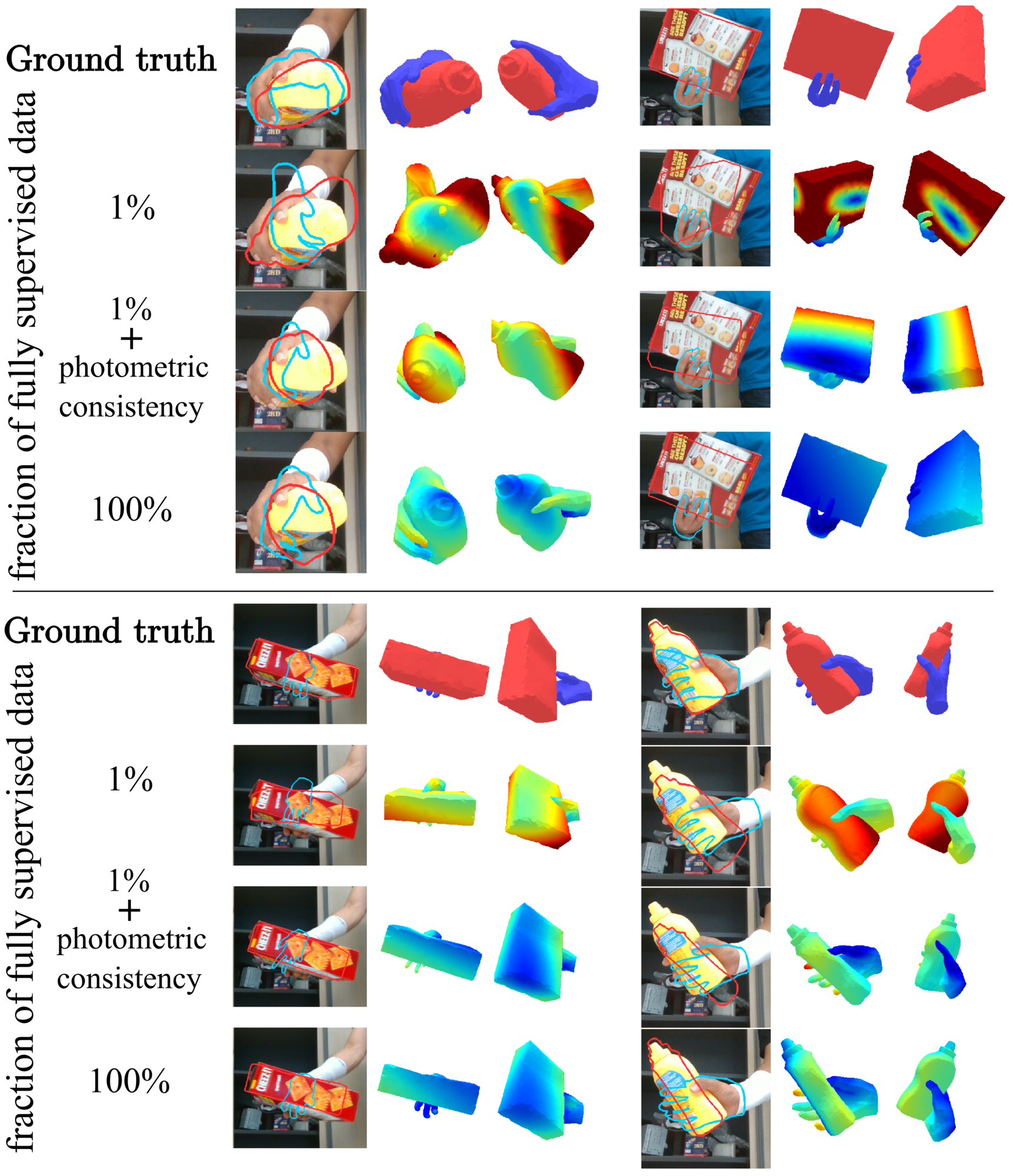}
    \end{center}
\caption{Predicted reconstructions for images from HO-3D. While rotation errors around axis parallel to the camera plane are not corrected and are sometimes even introduced by the photometric consistency loss, we observe qualitative improvement in the 2D reprojection of the predicted meshes on the image plane.}
    \mbox{}\vspace{-0.8cm}\\
    \label{fig:quali_ho-3d}
\end{figure}

\begin{table}
	\begin{center}
		\begin{tabular}{|l|c|c|}
			\hline
			& Hand error (mm)& Object error (mm) \\
			\hline\hline
			Hand only & 15.7 & - \\
			Object only & - & 21.8 \\
			Hand + Object & 18.0 & 22.3 \\
			\hline
		\end{tabular}
	\end{center}
	\caption{We compare training for hand and object pose estimation jointly and separately on FPHAB~\cite{hernando2018cvpr} and find that the encoder can be shared at a minor performance cost in hand and object pose accuracy.}
        \mbox{}\vspace{-0.8cm}\\
	\label{table:sharenc}
            \mbox{}\vspace{-1.2cm}\\
\end{table}

\begin{figure}[t]
\begin{center}
   \includegraphics[width=1\linewidth]{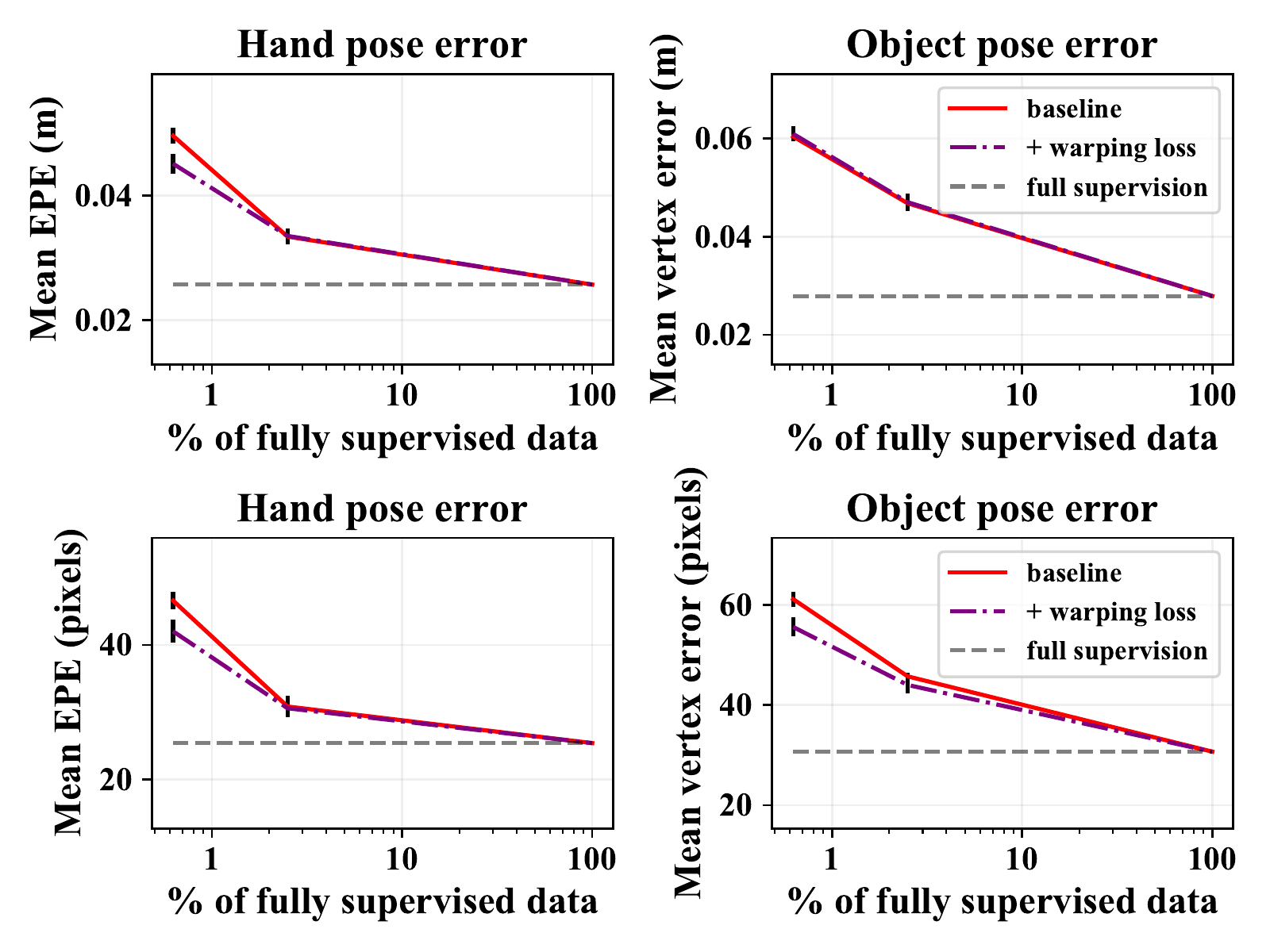}
\end{center}
    \mbox{}\vspace{-1.0cm}\\
   \caption{Effect of using photometric-consistency self-supervision when only a fraction of frames are fully annotated on HO-3D.
   We report average values and standard deviations over 5 different runs.}
    \mbox{}\vspace{-0.8cm}\\
\label{fig:photoconsistency_ho3d}
\end{figure}

\begin{figure}[t]
\begin{center}
   \includegraphics[width=1\linewidth]{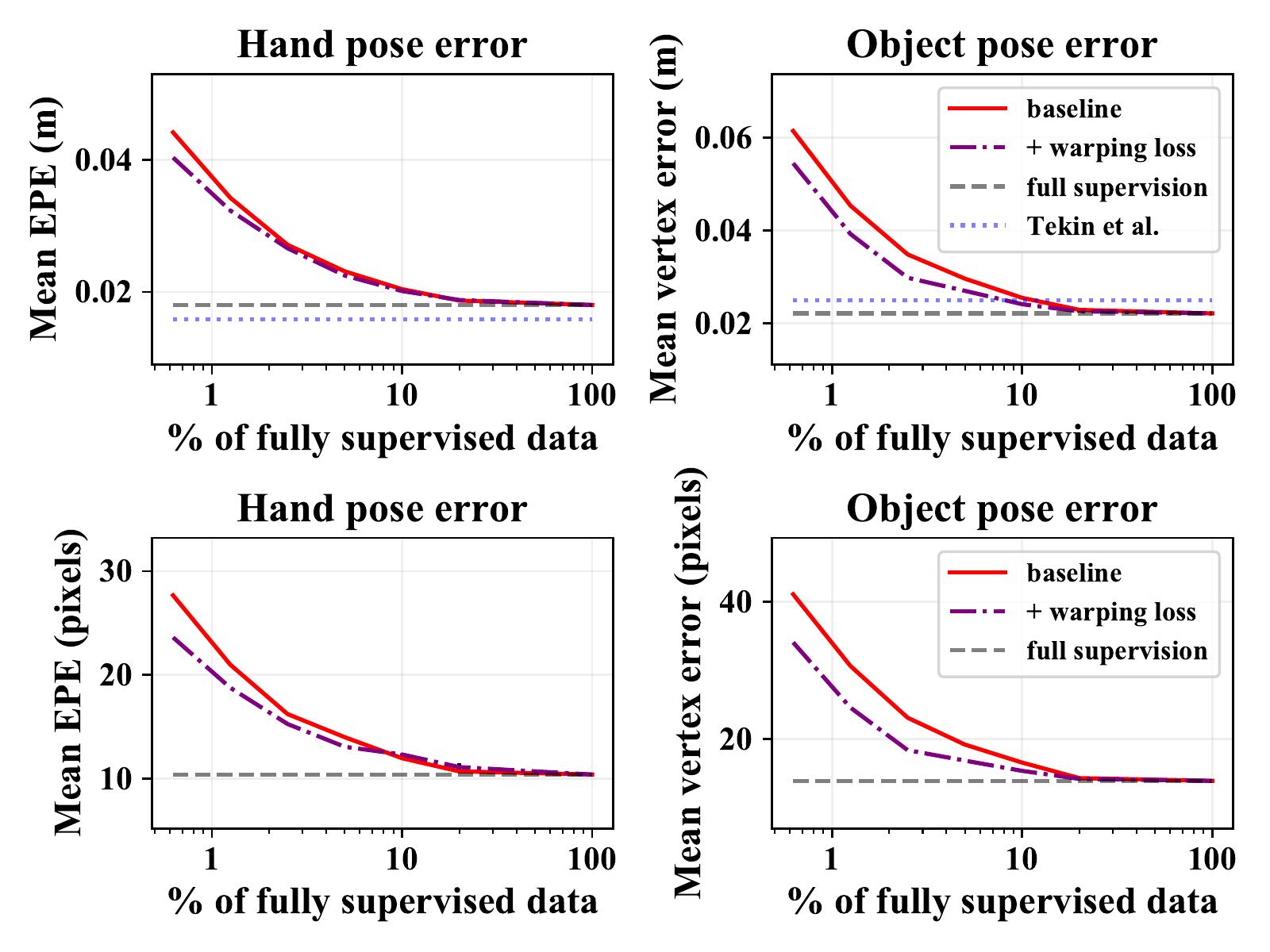}
\end{center}
    \mbox{}\vspace{-1.0cm}\\
   \caption{We observe consistent quantitative improvements from the photometric consistency loss as the percentage of fully supervised frames decreases below $10\% $ for both hands and objects.}
       \mbox{}\vspace{-0.8cm}\\
\label{fig:photoconsistency_fphab}
\end{figure}

\mbox{}\vspace{-0.6cm}\\

\textbf{Photometric supervision on video.}
We now validate the efficiency of our self-supervised dense hand-object reconstruction approach when ground-truth data availability is limited.
We pretrain several models on a fraction of the data by sampling frames uniformly in each sequence.
We sample a number of frames to reach the desired ratio of annotated frames in each training video sequence, starting from the first frame.
We then continue training with photometric consistency as an additional loss, while maintaining the full supervision on the sparsely annotated frames. Additional implementation and training details are discussed in Appendix~\ref{sec:implemdetails}. 
In order to single out the effect of the additional consistency term and factor out potential benefits from a longer training time, we continue training a reference model with the full supervision on the sparse keyframes for comparison.
We experiment with various regimes of data scarcity, progressively decreasing the percentage of annotated keyframes from $50$ to less than $1$\%.

\begin{figure*}[t]
\mbox{}\vspace{-.4cm}\\
\begin{center}
   \includegraphics[width=0.95\linewidth]{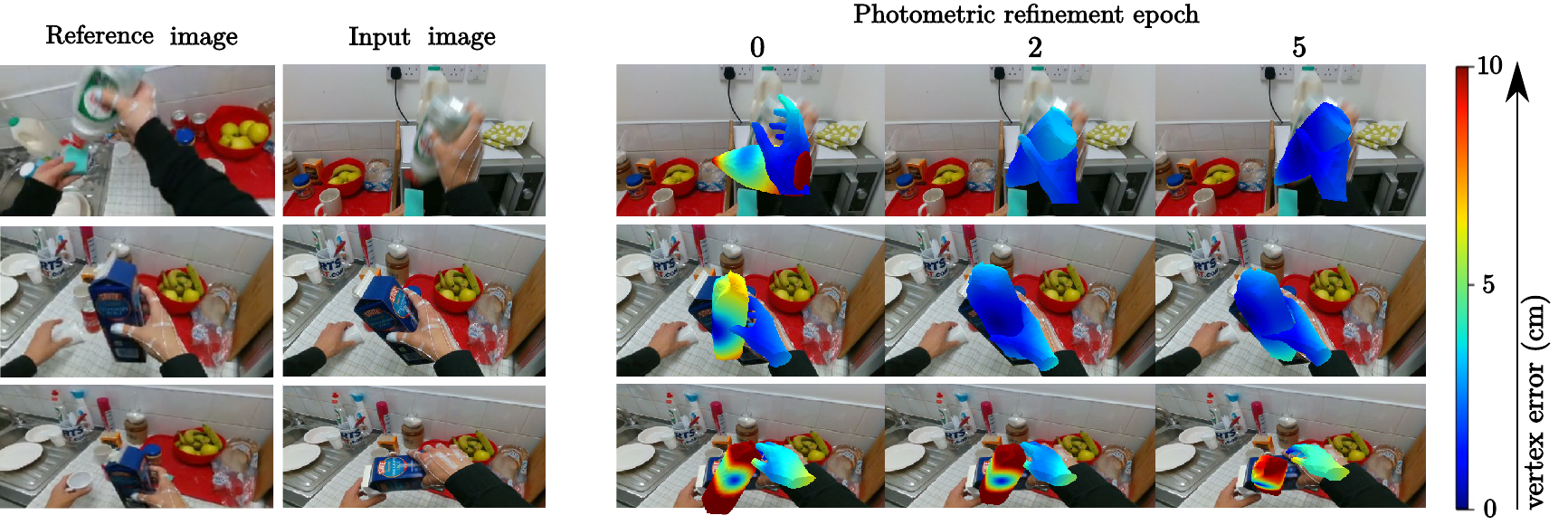}
\end{center}
   \caption{
   Progressive pose refinement over training samples, even in the presence of large motion and inaccurate initialization. 
   In extreme cases (last row), the model cannot recover. %
   }
\mbox{}\vspace{-0.85cm}\\
\label{fig:large_motion}
\end{figure*}

We report our results in Fig.~\ref{fig:photoconsistency_fphab} for FPHAB and in Fig.~\ref{fig:photoconsistency_ho3d} for HO-3D. 
We observe that only $20\%$ of the frames are necessary to reach the densely supervised performance on the FPHAB dataset, which can be explained by the correlated nature between neighboring frames. However, as we further decrease the fraction of annotated data, the generalization error significantly decreases.  
We demonstrate that our self-supervised learning strategy significantly improves the pose estimation accuracy in the low data regime when only a few percent of the actual dataset size are annotated and reduces the rigid reliance on large labeled datasets for hand-object reconstruction.
Although the similarity between the reference and consistency-supervised frames decreases
as the supervision across video becomes more sparse and the average distance to the reference frame increases, resulting in larger appearance changes, we observe that the benefits from our additional photometric consistency is most noticeable for both hands and objects as scarcity of fully annotated data increases.
When using less than one percent of the training data with full supervision, we observe an absolute average improvement of 7 pixels for objects and 4 pixels for hands, reducing the gap between the sparsely and fully supervised setting by respectively $25$ and $23\%$ (see Fig.~\ref{fig:photoconsistency_fphab}).
While on HO-3D the pixel-level improvements on objects do not translate to better 3D reconstruction scores for the object (see Fig.~\ref{fig:photoconsistency_ho3d}), on FPHAB, the highest relative improvement is observed for object poses when fully supervising $2.5\%$ of the data. In this setup, the $4.7$ reduction in the average pixel error corresponds to a reduction of the error by $51\%$ and results in a reduction by $40\%$ in the 3D $mm$ error.
We qualitatively investigate the modes of improvement and failure from introducing the additional photometric consistency loss in Fig.~\ref{fig:quali_fhb} and Fig.~\ref{fig:quali_ho-3d}.

As our method relies on photometric consistency for supervision, it is susceptible to fail when the photometric consistency assumption is infringed, which can occur for instance in cases of fast motions or illumination changes.
However, our method has the potential to provide meaningful supervision in cases where large motions occur between the reference and target frames, as long as the photometric consistency hypothesis holds.
We observe that in  most cases,  our  baseline  provides  reasonable  initial  pose  estimates on unannotated frames, which allows the photometric loss to provide informative gradients. In Fig.~\ref{fig:large_motion}, we show examples of successful and failed pose refinements on training samples from the FPHAB dataset supervised by our loss.
Our model is able to improve pose estimations in challenging cases,  where the initial prediction is inaccurate and there are large motions with respect to the reference frame.

\section{Conclusion}

In this paper, we propose a new method for dense 3D reconstruction of hands and objects from monocular color images. 
We further present a sparsely supervised learning approach leveraging photo-consistency between sparsely supervised frames.
We demonstrated that our approach achieves high accuracy for hand and object pose estimation and successfully leverages similarities between sparsely annotated and unannotated neighboring frames to provide additional supervision.
Future work will explore additional self-supervised 3D interpenetration and scene interaction constraints for hand-object reconstruction.
Our framework is general and can be extended to incorporate the full 3D human body along with the environment surfaces, which we intend to explore to achieve a full human-centric scene understanding.

\paragraph{Acknowledgments.}
\footnotesize
This work was supported by the MSR-Inria joint lab and the French government under management of Agence Nationale de la Recherche as part of the "Investissements d'avenir" program, reference ANR-19-P3IA-0001 (PRAIRIE 3IA Institute). Yana Hasson thanks Mihai Dusmanu, Yann Labbé and Thomas Eboli for helpful discussions.

{\small
\bibliographystyle{ieee_fullname}
\bibliography{egbib}

\begin{thebibliography}{10}\itemsep=-1pt

\bibitem{Arnab_CVPR_2019}
Anurag* Arnab, Carl* Doersch, and Andrew Zisserman.
\newblock Exploiting temporal context for 3d human pose estimation in the wild.
\newblock In {\em The IEEE Conference on Computer Vision and Pattern
  Recognition (CVPR)}, 2019.

\bibitem{bertasius2019neurips}
Gedas Bertasius, Christoph Feichtenhofer, Du Tran, Jianbo Shi, and Lorenzo
  Torresani.
\newblock Learning temporal pose estimation from sparsely-labeled videos.
\newblock In {\em Advances in Neural Information Processing Systems}, 2019.

\bibitem{Boukhayma_2019_CVPR}
Adnane Boukhayma, Rodrigo~de Bem, and Philip~H.S. Torr.
\newblock 3d hand shape and pose from images in the wild.
\newblock In {\em The IEEE Conference on Computer Vision and Pattern
  Recognition (CVPR)}, 2019.

\bibitem{Brickwedde_2019_ICCV}
Fabian Brickwedde, Steffen Abraham, and Rudolf Mester.
\newblock Mono-sf: Multi-view geometry meets single-view depth for monocular
  scene flow estimation of dynamic traffic scenes.
\newblock In {\em The IEEE International Conference on Computer Vision (ICCV)},
  October 2019.

\bibitem{cai2018_weakly}
Yujun Cai, Liuhao Ge, Jianfei Cai, and Junsong Yuan.
\newblock Weakly-supervised {3D} hand pose estimation from monocular {RGB}
  images.
\newblock In {\em The European Conference on Computer Vision (ECCV)}, 2018.

\bibitem{hernando2018cvpr}
Guillermo Garcia-Hernando, Shanxin Yuan, Seungryul Baek, and Tae-Kyun Kim.
\newblock First-person hand action benchmark with {RGB-D} videos and {3D} hand
  pose annotations.
\newblock In {\em The IEEE Conference on Computer Vision and Pattern
  Recognition (CVPR)}, 2018.

\bibitem{Ge2019}
Liuhao Ge, Zhou Ren, Yuncheng Li, Zehao Xue, Yingying Wang, Jianfei Cai, and
  Junsong Yuan.
\newblock 3d hand shape and pose estimation from a single rgb image.
\newblock In {\em The IEEE Conference on Computer Vision and Pattern
  Recognition (CVPR)}, 2019.

\bibitem{monodepth17}
Cl{\'{e}}ment Godard, Oisin {Mac Aodha}, and Gabriel~J. Brostow.
\newblock Unsupervised monocular depth estimation with left-right consistency.
\newblock In {\em The IEEE Conference on Computer Vision and Pattern
  Recognition (CVPR)}, 2017.

\bibitem{densepose}
Riza~Alp Guler, Natalia Neverova, and Iasonas Kokkinos.
\newblock {DensePose}: {D}ense human pose estimation in the wild.
\newblock In {\em The IEEE Conference on Computer Vision and Pattern
  Recognition (CVPR)}, 2018.

\bibitem{hampali2019ho3d}
Shreyas Hampali, Markus Oberweger, Mahdi Rad, and Vincent Lepetit.
\newblock Ho-3d: A multi-user, multi-object dataset for joint 3d hand-object
  pose estimation.
\newblock In {\em arXiv Preprint 1907.01481v1}, 2019.

\bibitem{hampali2019cvpr}
Shreyas Hampali, Markus Oberweger, Mahdi Rad, and Vincent Lepetit.
\newblock Honnotate: A method for 3d annotation of hand and objects poses.
\newblock In {\em The IEEE Conference on Computer Vision and Pattern
  Recognition (CVPR)}, 2020.

\bibitem{hasson19_obman}
Yana Hasson, G{\"u}l Varol, Dimitris Tzionas, Igor Kalevatykh, Michael~J.
  Black, Ivan Laptev, and Cordelia Schmid.
\newblock Learning joint reconstruction of hands and manipulated objects.
\newblock In {\em The IEEE Conference on Computer Vision and Pattern
  Recognition (CVPR)}, 2019.

\bibitem{He2015}
Kaiming He, Xiangyu Zhang, Shaoqing Ren, and Jian Sun.
\newblock Deep residual learning for image recognition.
\newblock In {\em The IEEE Conference on Computer Vision and Pattern
  Recognition (CVPR)}, 2015.

\bibitem{Hur_2017_ICCV}
Junhwa Hur and Stefan Roth.
\newblock Mirror{F}low: {E}xploiting symmetries in joint optical flow and
  occlusion estimation.
\newblock In {\em The IEEE International Conference on Computer Vision (ICCV)},
  2017.

\bibitem{flownet2}
Eddy Ilg, Nikolaus Mayer, Tonmoy Saikia, Margret Keuper, Alexey Dosovitskiy,
  and Thomas Brox.
\newblock Flownet 2.0: Evolution of optical flow estimation with deep networks.
\newblock In {\em The IEEE Conference on Computer Vision and Pattern
  Recognition (CVPR)}, 2017.

\bibitem{batchnorm15}
Sergey Ioffe and Christian Szegedy.
\newblock Batch normalization: Accelerating deep network training by reducing
  internal covariate shift.
\newblock In {\em International Conference on Machine Learning (ICML)}, 2015.

\bibitem{iqbal2018ECCV}
Umar Iqbal, Pavlo Molchanov, Thomas Breuel, Juergen Gall, and Jan Kautz.
\newblock Hand pose estimation via latent 2.5d heatmap regression.
\newblock In {\em The European Conference on Computer Vision (ECCV)}, 2018.

\bibitem{Totalcap2018}
Hanbyul Joo, Tomas Simon, and Yaser Sheikh.
\newblock Total capture: {A} 3d deformation model for tracking faces, hands,
  and bodies.
\newblock In {\em The IEEE Conference on Computer Vision and Pattern
  Recognition (CVPR)}, 2018.

\bibitem{hmrKanazawa18}
Angjoo Kanazawa, Michael~J. Black, David~W. Jacobs, and Jitendra Malik.
\newblock End-to-end recovery of human shape and pose.
\newblock In {\em The IEEE Conference on Computer Vision and Pattern
  Recognition (CVPR)}, 2018.

\bibitem{kato2018renderer}
Hiroharu Kato, Yoshitaka Ushiku, and Tatsuya Harada.
\newblock Neural {3D} mesh renderer.
\newblock In {\em The IEEE Conference on Computer Vision and Pattern
  Recognition (CVPR)}, 2018.

\bibitem{Kehl2017}
Wadim Kehl, Fabian Manhardt, Federico Tombari, Slobodan Ilic, and Nassir Navab.
\newblock Ssd-6d: Making rgb-based 3d detection and 6d pose estimation great
  again.
\newblock In {\em The IEEE International Conference on Computer Vision (ICCV)},
  2017.

\bibitem{adam2014}
Diederik~P. Kingma and Jimmy Ba.
\newblock Adam: {A} method for stochastic optimization.
\newblock {\em International Conference on Learning Representations}, 2014.

\bibitem{pnp}
Vincent Lepetit, Francesc Moreno-Noguer, and Pascal Fua.
\newblock {EPnP: An} accurate {O(n)} solution to the {PnP} problem.
\newblock {\em International Journal of Computer Vision (IJCV)}, 2009.

\bibitem{li2018deepim}
Yi Li, Gu Wang, Xiangyang Ji, Yu Xiang, and Dieter Fox.
\newblock Deepim: Deep iterative matching for 6d pose estimation.
\newblock In {\em The European Conference on Computer Vision (ECCV)}, 2018.

\bibitem{SMPL:2015}
Matthew Loper, Naureen Mahmood, Javier Romero, Gerard Pons-Moll, and Michael~J.
  Black.
\newblock {SMPL}: A skinned multi-person linear model.
\newblock {\em ACM Transactions on Graphics, (Proc. SIGGRAPH Asia)}, 2015.

\bibitem{GANeratedHands_CVPR2018}
Franziska Mueller, Florian Bernard, Oleksandr Sotnychenko, Dushyant Mehta,
  Srinath Sridhar, Dan Casas, and Christian Theobalt.
\newblock {GANerated} hands for real-time {3D} hand tracking from monocular
  {RGB}.
\newblock In {\em The IEEE Conference on Computer Vision and Pattern
  Recognition (CVPR)}, 2018.

\bibitem{mueller2019}
Franziska Mueller, Micah Davis, Florian Bernard, Oleksandr Sotnychenko, Mickeal
  Verschoor, Miguel~A. Otaduy, Dan Casas, and Christian Theobalt.
\newblock {Real-time Pose and Shape Reconstruction of Two Interacting Hands
  With a Single Depth Camera}.
\newblock {\em ACM Transactions on Graphics (TOG)}.

\bibitem{Mueller2017ICCV}
Franziska Mueller, Dushyant Mehta, Oleksandr Sotnychenko, Srinath Sridhar, Dan
  Casas, and Christian Theobalt.
\newblock Real-time hand tracking under occlusion from an egocentric {RGB-D}
  sensor.
\newblock In {\em The IEEE International Conference on Computer Vision (ICCV)},
  2017.

\bibitem{neverova19slim}
Natalia Neverova, James Thewlis, Riza~Alp Guler, Iasonas Kokkinos, and Andrea
  Vedaldi.
\newblock Slim densepose: Thrifty learning from sparse annotations and motion
  cues.
\newblock In {\em The IEEE Conference on Computer Vision and Pattern
  Recognition (CVPR)}, 2019.

\bibitem{pix2pose}
Kiru Park, Timothy Patten, and Markus Vincze.
\newblock Pix2{P}ose: {P}ixel-wise coordinate regression of objects for {6D}
  pose estimation.
\newblock In {\em The IEEE International Conference on Computer Vision (ICCV)},
  2019.

\bibitem{paszke2017automatic}
Adam Paszke, Sam Gross, Soumith Chintala, Gregory Chanan, Edward Yang, Zachary
  DeVito, Zeming Lin, Alban Desmaison, Luca Antiga, and Adam Lerer.
\newblock Automatic differentiation in {PyTorch}.
\newblock In {\em Advances in Neural Information Processing Systems Autodiff
  Workshop}, 2017.

\bibitem{Pavlakos_2019_ICCV}
Georgios Pavlakos, Nikos Kolotouros, and Kostas Daniilidis.
\newblock Texturepose: Supervising human mesh estimation with texture
  consistency.
\newblock In {\em The IEEE International Conference on Computer Vision (ICCV)},
  2019.

\bibitem{pavlakos2018humanshape}
Georgios Pavlakos, Luyang Zhu, Xiaowei Zhou, and Kostas Daniilidis.
\newblock Learning to estimate 3{D} human pose and shape from a single color
  image.
\newblock In {\em The IEEE Conference on Computer Vision and Pattern
  Recognition (CVPR)}, 2018.

\bibitem{pavllo:videopose3d:2019}
Dario Pavllo, Christoph Feichtenhofer, David Grangier, and Michael Auli.
\newblock 3d human pose estimation in video with temporal convolutions and
  semi-supervised training.
\newblock In {\em The IEEE Conference on Computer Vision and Pattern
  Recognition (CVPR)}, 2019.

\bibitem{pfister2015flowing}
Tomas Pfister, James Charles, and Andrew Zisserman.
\newblock Flowing convnets for human pose estimation in videos.
\newblock In {\em The IEEE International Conference on Computer Vision (ICCV)},
  2015.

\bibitem{Rad2017}
Mahdi Rad and Vincent Lepetit.
\newblock {BB8: A} scalable, accurate, robust to partial occlusion method for
  predicting the {3D} poses of challenging objects without using depth.
\newblock In {\em The IEEE International Conference on Computer Vision (ICCV)},
  2017.

\bibitem{Rhodin_eccv2018}
Helge Rhodin, Mathieu Salzmann, and Pascal Fua.
\newblock Unsupervised geometry-aware representation learning for {3D} human
  pose estimation.
\newblock In {\em The European Conference on Computer Vision (ECCV)}, 2018.

\bibitem{Rhodin_cvpr2018}
Helge Rhodin, Jörg Spörri, Isinsu Katircioglu, Victor Constantin, Frédéric
  Meyer, Erich Müller, Mathieu Salzmann, and Pascal Fua.
\newblock Learning monocular {3D} human pose estimation from multi-view images.
\newblock In {\em The IEEE Conference on Computer Vision and Pattern
  Recognition (CVPR)}, 2018.

\bibitem{RogezSR15}
Gr{\'{e}}gory Rogez, James Steven~Supancic III, and Deva Ramanan.
\newblock Understanding everyday hands in action from {RGB-D} images.
\newblock In {\em The IEEE International Conference on Computer Vision (ICCV)},
  2015.

\bibitem{romero2010b}
Javier Romero, Hedvig Kjellstr{\"o}m, and Danica Kragic.
\newblock Hands in action: real-time 3{D} reconstruction of hands in
  interaction with objects.
\newblock 2010.

\bibitem{MANO:SIGGRAPHASIA:2017}
Javier Romero, Dimitrios Tzionas, and Michael~J. Black.
\newblock Embodied hands: Modeling and capturing hands and bodies together.
\newblock {\em ACM Transactions on Graphics, (Proc. SIGGRAPH Asia)}, 2017.

\bibitem{ILSVRC15}
Olga Russakovsky, Jia Deng, Hao Su, Jonathan Krause, Sanjeev Satheesh, Sean Ma,
  Zhiheng Huang, Andrej Karpathy, Aditya Khosla, Michael Bernstein,
  Alexander~C. Berg, and Li Fei-Fei.
\newblock {ImageNet Large Scale Visual Recognition Challenge}.
\newblock {\em International Journal of Computer Vision (IJCV)}, 2015.

\bibitem{Simon_2017_CVPR}
Tomas Simon, Hanbyul Joo, Iain Matthews, and Yaser Sheikh.
\newblock Hand keypoint detection in single images using multiview
  bootstrapping.
\newblock In {\em The IEEE Conference on Computer Vision and Pattern
  Recognition (CVPR)}, 2017.

\bibitem{spurr2018cvpr}
Adrian Spurr, Jie Song, Seonwook Park, and Otmar Hilliges.
\newblock Cross-modal deep variational hand pose estimation.
\newblock In {\em The IEEE Conference on Computer Vision and Pattern
  Recognition (CVPR)}, 2018.

\bibitem{RealtimeHO_ECCV2016}
Srinath Sridhar, Franziska Mueller, Michael Zollhoefer, Dan Casas, Antti
  Oulasvirta, and Christian Theobalt.
\newblock Real-time joint tracking of a hand manipulating an object from rgb-d
  input.
\newblock In {\em The European Conference on Computer Vision (ECCV)}, 2016.

\bibitem{sundermeyer2018}
Martin Sundermeyer, Marton Zoltan-Csaba, Maximilian Durner, Manuel Brucker, and
  Rudolph Triebel.
\newblock Implicit {3D} orientation learning for {6D} object detection from
  {RGB} images.
\newblock In {\em The European Conference on Computer Vision (ECCV)}, 2018.

\bibitem{tekin19_handplusobject}
Bugra Tekin, Federica Bogo, and Marc Pollefeys.
\newblock {H+O}: Unified egocentric recognition of {3D} hand-object poses and
  interactions.
\newblock In {\em The IEEE Conference on Computer Vision and Pattern
  Recognition (CVPR)}, 2019.

\bibitem{tekin18}
Bugra Tekin, Sudipta Sinha, and Pascal Fua.
\newblock Real-time seamless single shot {6D} object pose prediction.
\newblock In {\em The IEEE Conference on Computer Vision and Pattern
  Recognition (CVPR)}, 2018.

\bibitem{tung2017self}
Hsiao-Yu Tung, Hsiao-Wei Tung, Ersin Yumer, and Katerina Fragkiadaki.
\newblock Self-supervised learning of motion capture.
\newblock In {\em Advances in Neural Information Processing Systems}, 2017.

\bibitem{Tzionas:IJCV:2016}
Dimitrios Tzionas, Luca Ballan, Abhilash Srikantha, Pablo Aponte, Marc
  Pollefeys, and Juergen Gall.
\newblock Capturing hands in action using discriminative salient points and
  physics simulation.
\newblock {\em International Journal of Computer Vision (IJCV)}, 2016.

\bibitem{xiang2017posecnn}
Yu Xiang, Tanner Schmidt, Venkatraman Narayanan, and Dieter Fox.
\newblock Posecnn: A convolutional neural network for 6d object pose estimation
  in cluttered scenes.
\newblock {\em Robotics: Science and Systems (RSS)}, 2018.

\bibitem{yang2019}
Linlin Yang and Angela Yao.
\newblock Disentangling latent hands for image synthesis and pose estimation.
\newblock In {\em The IEEE Conference on Computer Vision and Pattern
  Recognition (CVPR)}, 2019.

\bibitem{Zhang_2019_ICCV}
Xiong Zhang, Qiang Li, Hong Mo, Wenbo Zhang, and Wen Zheng.
\newblock End-to-end hand mesh recovery from a monocular {RGB} image.
\newblock In {\em The IEEE International Conference on Computer Vision (ICCV)},
  2019.

\bibitem{zhoucvpr2017}
Tinghui Zhou, Matthew Brown, Noah Snavely, and David Lowe.
\newblock Unsupervised learning of depth and ego-motion from video.
\newblock In {\em The IEEE Conference on Computer Vision and Pattern
  Recognition (CVPR)}, 2017.

\bibitem{zb2017hand}
Christian Zimmermann and Thomas Brox.
\newblock Learning to estimate 3d hand pose from single rgb images.
\newblock In {\em The IEEE International Conference on Computer Vision (ICCV)},
  2017.

\bibitem{Zimmermann_2019_ICCV}
Christian Zimmermann, Duygu Ceylan, Jimei Yang, Bryan Russell, Max Argus, and
  Thomas Brox.
\newblock Freihand: A dataset for markerless capture of hand pose and shape
  from single rgb images.
\newblock In {\em The IEEE International Conference on Computer Vision (ICCV)},
  2019.

\end{thebibliography}
}

\clearpage \newpage
\appendix

\section*{Appendix}
Our main paper described a method for joint reconstruction of hands and objects, and proposed to leverage photometric consistency as an additional source of supervision in scenarios where ground truth is scarce.
We provide additional details on the implementation in Section~\ref{sec:implemdetails}, and describe the used training and test splits on the HO-3D dataset~\cite{hampali2019ho3d} in Section~\ref{sec:ho3dsub}. In Section~\ref{sec:cyclic}, we detail the cyclic consistency check that allows us to compute the valid mask for the photometric consistency loss. Section~\ref{sec:skeletonadapt} provides additional insights on the effect of using the skeleton adaptation layer.

\section{Implementation details}\label{sec:implemdetails}

\paragraph{Architecture.} We extract image features from the last layer of ResNet18~\cite{He2015} before softmax.
We regress in separate branches $6$ parameters for the global object translation and rotation, $3$ parameters for the global hand translation, and $28$ MANO parameters which account for global hand rotation, articulated pose and shape deformation.
The details of each branch are presented in Table~\ref{tab:architecture}.

\begin{table}[b]
	\centering
	\footnotesize{
		\vspace{-2mm}
		\begin{tabular}{c c c c}
			\toprule
			\multirowcell{2}{Branch} & \multirowcell{2}{Input \\ shape} & \multirowcell{2}{Output \\ shape} & \multirowcell{2}{ReLU} \\
			& & & \\ \midrule
			\texttt{Object pose}  & 512 &  256 & \checkmark  \\
			\texttt{regressor}    & 256 & 6 &  \\ \midrule
			\texttt{Hand translation} & 512 &  256 & \checkmark \\
			\texttt{regressor}  & 256 & 3 & \\ \midrule
			\texttt{Hand pose} & 512 &  512 & \checkmark \\
			\texttt{and shape}  & 512 & 512 & \checkmark \\
			\texttt{regressor}  & 512 & 28 & \\
			\bottomrule
			& & & \\
		\end{tabular}
	}
	\caption{\small {\bf Architecture of the Hand and Object parameter regression branches.} We use fully connected linear layers to regress pose and shape parameters from the $512-$dimensional features.}
	\label{tab:architecture}
\end{table}

\paragraph{Training.}
All models are trained using the PyTorch~\cite{paszke2017automatic} framework.
We use the Adam~\cite{adam2014} optimizer with a learning rate of $5\cdot 10^{-5}$.
We initialize the weights of our network using the weights of a ResNet~\cite{He2015} trained on ImageNet~\cite{ILSVRC15}. 
We empirically observed improved stability during training when freezing the weights of the batch normalization~\cite{batchnorm15} layer to the weights initialized on ImageNet.

We pretrain the models on fractions of the data without the consistency loss. As an epoch contains fewer iterations when using a subset of the dataset, we observe that a larger number of epochs is needed to reach convergence for smaller fractions of training data. We later fine-tune our network with the consistency loss using a fixed number of 200 epochs.

\paragraph{Runtime.}
The forward pass runs in real time, at 34 frames per second on a Titan X GPU.

\section{HO-3D subset}\label{sec:ho3dsub}

In Sec. 4.3, we work with the subset of the dataset which was first released.
Out of the 68 sequences which have been released as the final version of the dataset, 15 have been made available as part of an earlier release.
Out of these, we select the 14 sequences that depict manipulation of two following objects: the mustard bottle and the cracker box.
The train sequences in this subset are the ones named SM2, SM3, SM4, SM5, MC4, MC6, SS1, SS2, SS3, SM2, MC1, MC5.
When experimenting with the photometric consistency, we use SM1 and MC2 as the two test sequences.
When comparing to the baseline of~\cite{hampali2019ho3d}, we use MC2 as the unique test sequence.
{}

\section{Cycle consistent visibility check}\label{sec:cyclic}

Our consistency check is similar to~\cite{neverova19slim,Hur_2017_ICCV}.

Following the notation of Sec.~3.1, let us denote the flow warping the estimated frame $I_{t_{ref}+k}$ into the reference one $I_{t_{ref}}$ by $W_{t_{ref}+k\rightarrow t_{ref}}$. Similarly, we compute a warping flow in the opposite direction, from the reference frame to the estimated one: $W_{t_{ref}\rightarrow t_{ref}+k}$.
Given the mask $M_{t_{ref}}$ obtained by projecting $V_{t_{ref}}$ on image space, we consider each pixel $p \in M_{t_{ref}+k}$.
We warp $p$ into the reference frame, and then back into the estimated one: $\tilde{p} = W_{t_{ref}+k\rightarrow t_{ref}}(W_{t_{ref}\rightarrow t_{ref}+k}(p))$.
If the distance between $p$ and $\tilde{p}$ is greater than 2 pixels, we do not apply our loss at this location. On FHB, when using 1\% of the data as reference frames, this check discards 3.3\% of $M_{t_{ref}+k}$ pixels.

\section{Skeleton Adaptation}\label{sec:skeletonadapt}

\begin{figure}[t]	

    \begin{center}
        \vspace{-0.1cm}
        \includegraphics[width=0.7\linewidth]{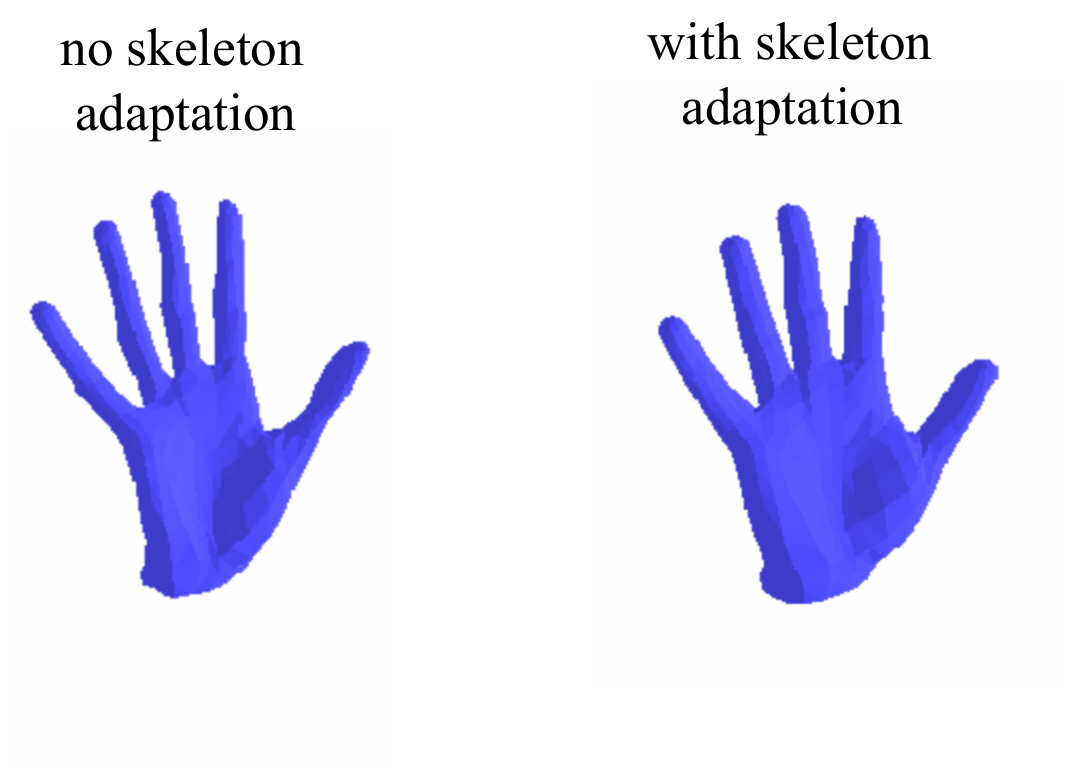} \\
        \vspace{-0.8cm}
        \footnotesize{\hspace{-0.2cm} \textbf{(a)} \hspace{3.0cm} \textbf{(b)}}
        \vspace{-0.4cm}
    \end{center}
    \caption{Predicted shape deformations in the \textbf{(a)} absence and \textbf{(b)} presence of the skeleton adaptation layer on the FPHAB dataset.}
    \label{fig:skeladapt}
    \vspace{-0.5cm}
\end{figure}

The defined locations for the joints do not exactly match each other for the FPHAB~\cite{hernando2018cvpr} dataset and the MANO~\cite{MANO:SIGGRAPHASIA:2017} hand model.
As shown in Table 2 of our main paper, we observe marginal improvements in the average joint predictions using our skeleton adaptation layer. This demonstrates that  MANO~\cite{MANO:SIGGRAPHASIA:2017} has already the ability to deform sufficiently to account for various skeleton conventions.
However, these deformations come at the expense of the realism of the reconstructed meshes, which undergo unnatural deformations in order to account for the displacements of the joints.
To demonstrate this effect, we train a model on the FPHAB~\cite{hernando2018cvpr} dataset, without the linear skeleton adaptation layer, and qualitatively compare the predicted hand meshes with and without skeleton adaptation.
We observe in Fig.~\ref{fig:skeladapt}(a) that, without skeleton adaptation, the fingers get unnaturally elongated to account for different definitions of the joint locations in FPHAB and MANO. As shown in Fig.~\ref{fig:skeladapt}(b), we are able to achieve higher realism for the reconstructed meshes using our skeleton adaptation layer.

\end{document}